# Voltage-Dependent Synaptic Plasticity (VDSP): Unsupervised probabilistic Hebbian plasticity rule based on neurons membrane potential

**Nikhil Garg[1,2,3,*], Ismael Balafrej[1,2,5], Terrence C. Stewart[6], Jean Michel Portal[4], Marc Bocquet[3], Damien Querlioz[7], Dominique Drouin[1,2], Jean Rouat[1,2,5], Yann Beilliard[1,2], Fabien Alibart[1,2,3,*]**

[1]Institut Interdisciplinaire d'Innovation Technologique (3IT), Université de Sherbrooke, Sherbrooke, Québec J1K 0A5, Canada

[2]Laboratoire Nanotechnologies Nanosystèmes (LN2) – CNRS UMI-3463, Université de Sherbrooke, Sherbrooke, Québec J1K 0A5, Canada

[3]Institute of Electronics, Microelectronics and Nanotechnology (IEMN), Université de Lille, 59650 Villeneuve d'Ascq, France

[4]Aix Marseille Univ, Université de Toulon, CNRS, IM2NP, Marseille, France

[5]NECOTIS Research Lab, Electrical and Computer Engineering Dep., Université de Sherbrooke, Quebec, Canada

[6]National Research Council Canada, University of Waterloo Collaboration Centre, Waterloo, ON, Canada

[7]Université Paris-Saclay, CNRS, Centre de Nanosciences et de Nanotechnologies, Palaiseau, France
* Correspondence:
Corresponding Author
Nikhil.Garg@Usherbrooke.ca
Fabien.Alibart@Usherbrooke.ca

**Abstract**

This study proposes voltage-dependent-synaptic plasticity (VDSP), a novel brain-inspired unsupervised local learning rule for the online implementation of Hebb's plasticity mechanism on neuromorphic hardware. The proposed VDSP learning rule updates the synaptic conductance on the spike of the postsynaptic neuron only, which reduces by a factor of two the number of updates with respect to standard spike timing dependent plasticity (STDP). This update is dependent on the membrane potential of the presynaptic neuron, which is readily available as part of neuron implementation and hence does not require additional memory for storage. Moreover, the update is also regularized on synaptic weight and prevents explosion or vanishing of weights on repeated stimulation. Rigorous mathematical analysis is performed to draw an equivalence between VDSP and STDP. To validate the system-level performance of VDSP, we train a single-layer spiking neural network (SNN) for the recognition of handwritten digits. We report 85.01 ± 0.76% (Mean±S.D.) accuracy for a network of 100 output neurons on the MNIST dataset. The performance improves when scaling the network size (89.93 ± 0.41% for 400 output neurons, 90.56 ± 0.27 for 500 neurons), which validates the applicability of the proposed learning rule for spatial pattern recognition tasks. Future work will consider more complicated tasks. Interestingly, the learning rule better adapts than STDP to the frequency of input signal and does not require hand-tuning of hyperparameters.

**Introduction**

The amount of data generated in our modern society is growing dramatically, and Artificial Intelligence (AI) appears as a highly effective option to process this information. However, AI still faces the major challenge of data labelling: machine learning algorithms associated with supervised learning can bring AI at human-level performance, but they require costly manual labelling of the datasets. A highly desirable alternative would be to deploy unsupervised learning strategies that do not require data pre-processing. Neuromorphic engineering and computing, which aims to replicate bio-realistic circuits and algorithms through a spike-based representation of data, relies heavily on such unsupervised learning strategies. Spike timing dependent plasticity (STDP) is a popular unsupervised learning rule used in this context, where the relative time difference between the pre- and post-synaptic neuron spikes defines synaptic plasticity (Lee et al., 2018; Brader, Senn and Fusi, 2007, Masquelier and Thorpe, 2007). STDP is a spiking version of the traditional Hebbian learning concept (Hebb, 1949; Bliss and Collingridge, 1993; Bi and Poo, 1998), where a synaptic connection is modified depending only on the local activity correlations between its presynaptic and postsynaptic neurons.

In addition to its intrinsic unsupervised characteristic, STDP is also very attractive due to the locality of its synaptic learning. Such a feature could dramatically reduce hardware constraints of SNN by avoiding complex data exchange at the network level. However, STDP retains a major challenge: it requires precise spike times/traces to be stored in memory and fetched at every update to the processor. In most implementations (Morrison, Diesmann and Gerstner, 2008; Song, Miller and Abbott, 2000), decaying spike traces are used to compute synaptic weight update, adding extra state variables to store and update. In digital neuromorphic systems (Diehl and Cook, 2014; Manoharan, Muralidhar and Kailath, 2020; Yousefzadeh et al., 2017; Lammie et al., 2019), implementing STDP comes with an added cost of memory requirement for storing spike times/traces for every neuron and energy expenditure for fetching these variables during weight update. For analog hardware implementation (Moriya et al., 2021; Friedmann et al., 2017; Grübl et al., 2020; Narasimman et al., 2016), circuit area and power are spent in storing spike traces on capacitors, thus raising design challenges. In-memory computing approaches have been strongly considered for STDP implementation to mitigate memory bandwidth requirements. The utilization of non-volatile memory-based synapses, or memristors, has been primarily considered (Ambrogio et al., 2016; Serrano-Gotarredona et al., 2013; Querlioz et al., 2011; Camunas-Mesa, Linares-Barranco and Serrano-Gotarredona, 2020). The seminal idea is to convert the time distance between pre- post-signals into a voltage applied across a single resistive memory element. The key advantage is to compute the STDP function directly on the memory device and to store the resulting synaptic weight permanently. This approach limits data movement and ensures the compactness of the hardware design (single memristor crosspoints may feature footprints below 100 nm). Further similar hardware propositions for STDP implementation have been discussed in the literature (Guo et al., 2019) and (Boybat et al., 2018). Nevertheless, in all these approaches, time-to-voltage conversion requires a complex pulse shape (pulse duration should be in the order of STDP window and pulse amplitude should reflect the shape of STDP function), thus requiring complex circuit overhead and limiting the energy benefit of low power memory devices.

Moreover, STDP has the constraint of a fixed time window. As STDP is a function of the spike time difference between a post and a presynaptic neurons, the time window is the region in which the spike time difference must fall to update the weight significantly. The region of the time windows must be optimized to the temporal dynamics of spike-based signals to achieve good performances with STDP. This latter point raises additional issues at both the computational level (i.e., how to choose the appropriate STDP time window) and hardware level (i.e., how to design circuits with this level of flexibility). In other words, the challenge for deploying unsupervised strategies in neuromorphic SNN is two-sided: the concept of STDP needs to be further developed to allow for robust learning performances, and hardware implementations opportunities need to be considered in the meantime to ensure large scale neuromorphic system development.

In this work, we propose Voltage-Dependent Synaptic Plasticity (VDSP), an alternative approach to STDP that addresses these two limitations of STDP: VDSP does not require a fixed scale of spike time difference to update the weights significantly and can be easily integrated on in-memory computing hardware by preserving local computing. Our approach uses the membrane potential of a pre-synaptic neuron instead of its spike timing to evaluate pre/post neurons correlation. For a Leaky Integrate-and-Fire (LIF) neuron (Abbott, 1999), membrane potential exhibits exponential decay and captures essential information about the neuron's spike time; intuitively, a high membrane potential could be associated with a neuron that is about to fire while low membrane potential reflects a neuron that has recently fired. A post-synaptic neuron spike event is used to trigger the weight update based on the state of the pre-synaptic neuron. The rule leads to a biologically coherent temporal difference. We validate the applicability of this unsupervised learning mechanism to solve a classic computer vision problem. We tested a network of spiking neurons connected by such synapses to perform recognition of handwritten digits and report similar performance to other single-layer networks trained in unsupervised fashion with the STDP learning rule. Remarkably, we show that the learning rule is resilient to the temporal dynamics of the input signal and eliminates the need to tune the hyperparameters for input signals of different frequency range. This approach could be implemented in neuromorphic hardware with little logic overhead, memory requirement and enable larger networks to be deployed in constrained hardware implementations.

Past studies have investigated the role of membrane potential in the plasticity of the mammalian cortex (Artola, Bröcher and Singer, 1990). The in-vivo voltage dependence of synaptic plasticity has been demonstrated in (Jedlicka, Benuskova, and Abraham, 2015. In (Clopath et al., 2010), bidirectional connectivity formulation in the cortex has been demonstrated as a resultant of voltage-dependent Hebbian-like plasticity. In (Diederich et al., 2018), a voltage-based Hebbian learning rule was used to program memristive synapses in a recurrent bidirectional network. A presynaptic spike led to a weight update dependent on the membrane potential of postsynaptic neurons. The membrane potential was compared with a threshold voltage. If the membrane potential exceeded this threshold, long-term potentiation (LTP) was applied by applying a fixed voltage pulse on the memristor, while, for low membrane potential, long-term depression (LTD) took place. However, in their case, the weight update is independent of the magnitude of the membrane potential, and hence the effect of precise spike time difference cannot be captured. Lastly, these past studies have never reported handwritten digit recognition and benchmark against STDP counterparts.

In the following sections, we first describe the spiking neuron model and investigate the relation between spike time and neuron membrane potential. Second, we describe the proposed plasticity algorithm, its rationale, and its governing equations. Third, the handwritten digit recognition task is described with SNN topology, neuron parameters and learning procedure. In the results section, we report the network's performance for handwritten digit recognition. Next, we demonstrate the frequency normalization capabilities of VDSP as opposed to STDP by trying widely different firing frequencies for the input neurons in the handwritten digit recognition task without adapting the parameters. Finally, the hyperparameter tuning and scalability of the network are discussed.

**Methods**

**Neuron modelling**

LIF neurons (Abbott, 1999) are simplified version of biological neurons, hence easy to simulate in an SNN simulator. This neuron model was used for the pre-synaptic neuron layers. The governing equation is

$$\tau_m \frac{dv}{dt} = -v + I + b \quad (1)$$

where $\tau_m$ is the membrane leak time constant, $v$ is the membrane potential, which leaks to resting potential ($v_{rest}$), $I$ is the injected current, and $b$ is a bias. Whenever the membrane potential exceeds a threshold potential ($v_{th}$), the neuron emits a spike. Then, it becomes insensitive to any input for the refractory period ($t_{ref}$) and the neuron potential is reset to voltage ($v_{reset}$).

An adaptation mechanism is added to the post neurons to prevent instability due to excessive firing. In the resulting adaptive leaky integrate-and-fire (ALIF) neuron, a second state variable is added. This state variable $n$ is increased by *inc_n* whenever a spike occurs, and the value of $n$ is subtracted from the input current. This causes the neuron to reduce its firing rate over time when submitted to strong input currents (Camera et al., 2004). The state variable $n$ decays by $\tau_n$:

$$\tau_n \frac{dn}{dt} = -n \quad\quad (2)$$

**Relation between spike time and membrane potential**

Hebbian-based STDP can be defined as the relation between $\Delta w \in \mathbb{R}$, the change in the conductance of a weight, and $\Delta t = t_{post} - t_{pre}$, the time interval between a presynaptic spike at time $t_{pre}$ and a postsynaptic spike at time $t_{post}$ with $\Delta t, t_{pre}, t_{post} \in \mathbb{R}^+$. This relation can be modelled as

$$\Delta w \propto \begin{cases} exp\left(\frac{-\Delta t}{\tau_{STDP}^+}\right), & t_{pre} < t_{post} \\ -exp\left(\frac{\Delta t}{\tau_{STDP}^-}\right), & \text{otherwise.} \end{cases} \quad (3)$$

with $\tau_{STDP}^{\pm}$ being the time constants for potentiation (+) and depression (-). This model is commonly computed during both the pre and postsynaptic neuron spikes, e.g., with the two traces model (Song, Miller and Abbott, 2000). For VDSP, we seek to compute a similar $\Delta w$, but as a function of only $v(t_{post})$, the membrane potential of a presynaptic neuron at the time of a postsynaptic spike.

Fortunately, when the presynaptic LIF neuron is only fed by a constant positive current $I \in \mathbb{R}^+$, the spiking dynamics can be predicted. Solving the presynaptic LIF neuron's differential equation for the membrane potential with no bias (eq. 1 with $b = 0$) during subthreshold behavior yields

$$v(t) = I + c \cdot \exp\left(-\frac{t}{\tau_m}\right), \qquad (4)$$

where $c$ is the integration constant. Solving eq. 4 for $t_{\text{pre}}$ and $t_{\text{post}}$ allows us to define a new relation for $t_{\text{post}} - t_{\text{pre}}$:

$$t_{\text{post}} - t_{\text{pre}} = \tau_m \ln\left(\frac{v(t_{\text{pre}}) - I}{v(t_{\text{post}}) - I}\right). \qquad (5)$$

with $v(t_{\text{pre}})$ and $v(t_{\text{post}})$ equal to the membrane potential of the presynaptic neuron at the moment of a presynaptic spike and postsynaptic spike respectively. Assuming $I$ is sufficient to make the presynaptic neuron spike in a finite amount of time, i.e., $I > v_{\text{th}}$, then $v(t_{\text{pre}} - \epsilon) = v_{\text{th}}$ and $v(t_{\text{pre}} + \epsilon) = v_{\text{reset}}$, with $\epsilon$ representing an infinitesimal number. Conceptually, $v(t_{\text{pre}} - \epsilon)$ represents a spike that is about to happen and $v(t_{\text{pre}} + \epsilon)$ a spike that has happened in the recent past, when there is no refractory period ($t_{ref} = 0$). Assuming $\epsilon \to 0$, we obtain: $\Delta t = \tau_m \ln\left(\frac{v_{\text{th}} - I}{v(t_{\text{post}}) - I}\right)$ if the presynaptic neuron is about to spike or $\Delta t = \tau_m \ln\left(\frac{v_{\text{reset}} - I}{v(t_{\text{post}}) - I}\right)$ if the presynaptic neuron recently spiked. To select between one of these values, we must obtain the smallest $\Delta t$, as to form a pair of $t_{\text{pre}}$ and $t_{\text{post}}$ that are closest in time. These two equations can be combined into:

$$|\Delta t| = \tau_m \cdot \min\left\{ \left|\ln\left(\frac{v_{\text{th}} - I}{v(t_{\text{post}}) - I}\right)\right|, \left|\ln\left(\frac{v_{\text{reset}} - I}{v(t_{\text{post}}) - I}\right)\right|\right\} \quad (6)$$

By using $\Delta t$ as a function of $v(t_{\text{post}})$ from equation 6, with $v_{\text{th}} = 1$, $v_{\text{reset}} = -1$ and knowing $v_{\text{reset}} \leq v(t_{\text{post}}) < v_{\text{th}}$, then equation 1 can be rearranged to:

$$\Delta w \propto \begin{cases} \left(\dfrac{v(t_{\text{post}}) - I}{-1 - I}\right)^{\frac{\tau_m}{\tau_{\text{STDP}}^+}}, & I - \sqrt{I^2 - 1} > v(t_{\text{post}}) \\[2ex] -\left(\dfrac{1 - I}{v(t_{\text{post}}) - I}\right)^{\frac{\tau_m}{\tau_{\text{STDP}}^-}}, & \text{otherwise.} \end{cases} \qquad (7)$$

This final result proves that, when the presynaptic neuron is driven by constant current, Hebbian STDP can be precisely modelled using only $v(t_{\text{post}})$, the membrane potential of a presynaptic neuron at the time of a postsynaptic spike. Note that such generalization cannot be done in the case of Poisson-like input signals. Fig. 1 (A-B) demonstrates experimentally the relation between the membrane potential and $|\Delta t|$ from equation 6. The condition $I - \sqrt{I^2 - 1} > v(t_{\text{post}})$ can be inferred from eq. 6, to select the minimal parameter, since $\min\{a, b\} = \begin{cases} a, & a \leq b \\ b, & \text{otherwise.} \end{cases}$

Moreover, as equation 6 shows, the neuron parameters, namely the membrane reset and threshold potentials, are implicitly used to calculate the potentiation and depression windows. For example, the condition $I - \sqrt{(I^2 - 1)} > v(t_{\text{post}})$ of equation 7 can be simplified to $v(t_{\text{post}}) < 0$ if $v_{\text{reset}} = \frac{I}{v_{\text{th}} - I}$ instead of $-1$. Both $v_{\text{th}}$ and $v_{\text{reset}}$ can be modified to tune the balance between potentiation and depression. Supplementary Figure S4 highlights the empirical effect of changing the value of $v_{th}$ and $v_{reset}$ on the $\Delta w = VDSP(\Delta t)$ window between two neurons with a fixed initial weight $w = 0.5$.

## Proposed plasticity algorithm

The proposed implementation of synaptic plasticity depends on the postsynaptic neuron spike time and the presynaptic neuron's membrane potential. This version of Hebbian plasticity in which the weight is updated on either postsynaptic or presynaptic spikes is also known as single spike synaptic plasticity (Serrano-Gotarredona et al., 2013). In real world applications, the presynaptic input current $I$ is often not known and not constant, which would be mandatory for reproducing STDP perfectly as demonstrated in eq. 7. The less information is known about the input current, the more our plasticity rule converge into a probabilistic model. A low membrane potential suggests that the presynaptic neuron has fired recently, leading to synaptic potentiation (Fig. 1(C)-(E)). A high presynaptic membrane potential suggests that the pre-synaptic neuron might fire shortly in the future and leads to depression (Fig. 1(F)- (H)). A different resting state potential and reset potential is essential to discriminate inactive neurons and neurons that spiked recently.

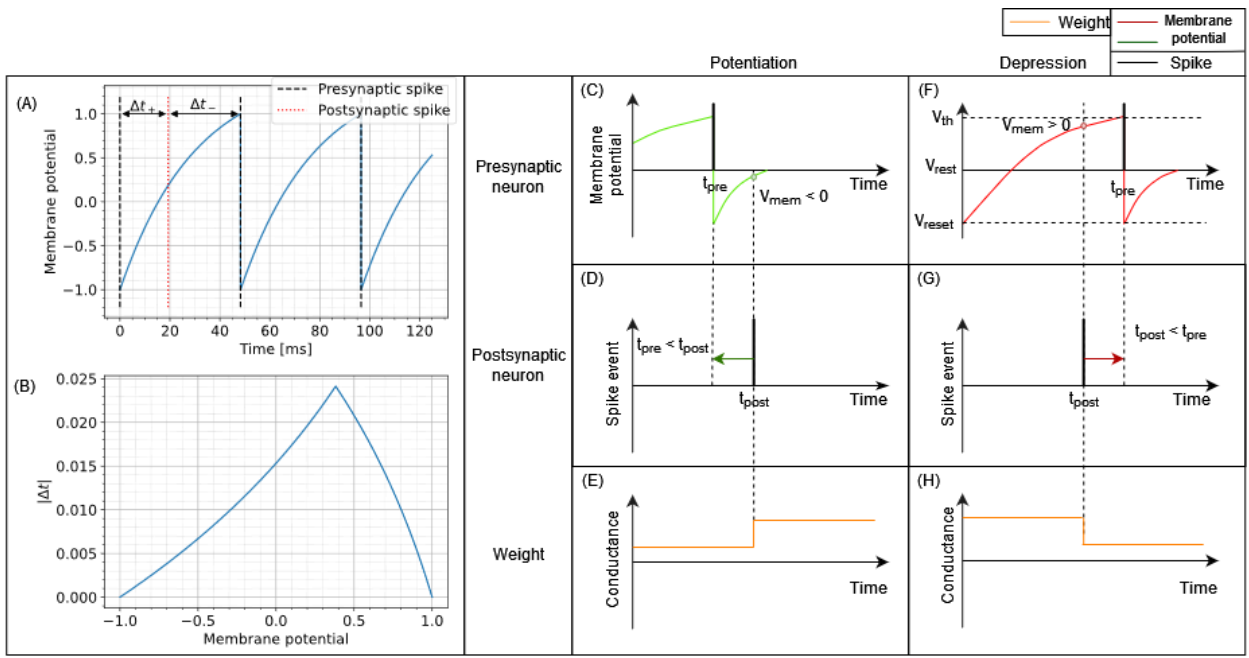

**Fig.1**: Schematic representation of the VDSP learning rule implemented between a pre- and postsynaptic spiking neuron. In (A), the membrane potential of a LIF neuron is shown evolving through time when fed with a constant current. In (B), the absolute time difference between the post and presynaptic spikes is computed analytically as a function of the membrane potential from (A). It is trivial, once the spike time difference is computed, to determine the STDP window as a function of membrane potential. Panels (C) and (F) show the spiking event of the presynaptic neuron (vertical black line) along with its membrane potential (coloured curve). Panels (D) and (G) show the spike event of the postsynaptic neuron. The weight update (panel (E) and (H)) happens whenever the post-synaptic neuron fires. The update is dependent on the membrane potential of pre-synaptic neuron. If the pre-synaptic neuron fired in the recent past ($t_{pre} < t_{post}$), the membrane potential of the presynaptic neuron is lesser than zero, and we observe potentiation of synaptic weight (panel (C)-(E)). Whereas if the pre-synaptic neuron is about to fire ($t_{post} < t_{pre}$), the membrane potential of the pre-synaptic neuron is greater than zero and we observe depression of synaptic weight (panel (F)-(H)).

Hebbian plasticity mechanisms can be grouped into additive or multiplicative types. In the additive versions of plasticity, the magnitude of weight update is independent of the current weight, but weight clipping must be implemented to restrict the values of weight between bounds (Brader et al., 2007). Although the weight is not present in weight change computation equation directly, the present weight must be fetched for applying clipping. In neurophysiology experiments (Rossum et al., 2000), it is also demonstrated that the weight update depends on the current synaptic weight in addition to the temporal correlation of spikes and is responsible for stable learning. The weight dependence is often referred to as multiplicative Hebbian learning as opposed to its additive counterpart and leads to stable learning and log-normal distribution of firing rates which are coherent with biological system recording (Teramae and Fukai, 2014).

VDSP relies on the multiplicative plasticity rule that considers the present weight value for computing the weight update magnitude. During potentiation, the weight update is proportional to ($W_{max}$-$W$), and during the depression phase, the weight update magnitude is proportional to $W$, where $W$ is the current weight, and $W_{max}$ is the maximum weight. Multiplicative weight dependence is a crucial feature of VDSP, and no hardbound is needed as typically used with additive plasticity rules. A detailed discussion is presented in the discussion section and Supplementary Fig. S2.

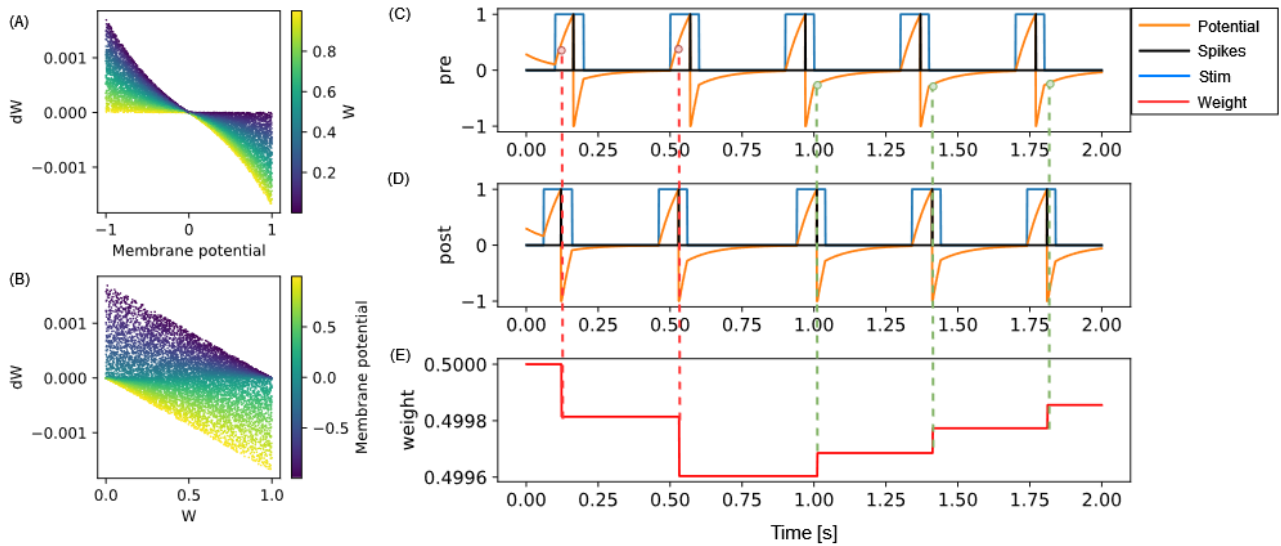

**Fig. 2**: (A) The weight update (dW) is plotted as a function of the membrane potential of pre-synaptic neuron, with the colour code representing the initial weight. (B) The dW is linearly dependent on (1-W) for potentiation and on (W) for depression. The learning rate is set to 0.001 in both (A) and (B). (C)-(E) A pair of pre-synaptic neuron and post-synaptic neuron is simulated along with their synaptic weight evolution. The weight update occurs at every post-synaptic neuron spike event and is negative if the pre-synaptic neuron membrane potential is greater than zero

(shown in red dotted lines). The weight update is positive (green dotted lines) if the pre-synaptic neuron voltage is lesser than zero.

The functional dependence of weight update on the membrane potential of the presynaptic neuron and the current synaptic weight is presented in Figures 2(A) and 2(B). The weight or synaptic conductance varies between zero and one. The weight update is modelled as

$$\Delta w = \begin{cases} \delta(t - t_{post})(w_{max} - w)(e^{-V_{pre}} - 1)lr, & V_{pre} < 0 \\ -\delta(t - t_{post})w(e^{V_{pre}} - 1)lr, & V_{pre} > 0 \end{cases} \quad (10)$$

where $dW$ is the change in weight, $V_{pre}$ is the membrane potential of the presynaptic neuron, $t_{post}$ is the time of postsynaptic neuron spike event, $W$ is the current weight of the synapse, $W_{max}$ is the maximum weight and is set to one, $t$ is the current time, and $lr$ is the learning rate.

To illustrate the weight update in the SNN simulator, a pair of neurons (Fig. 2(C) and 2(D)) were connected through a synapse (Fig. 2(E)) implementing the VDSP learning rule. The presynaptic and postsynaptic neurons were forced to spike at specific times. Potentiation and depression for $t_{post} > t_{pre}$ and $t_{post} < t_{pre}$ are shown with green and red dotted lines, respectively.

## MNIST Classification network

To benchmark the learning efficiency of the proposed learning rule for pattern recognition, we perform recognition of handwritten digits. One advantage of this task is that the weights of the trained networks can be interpreted to evaluate the network's learning. We use the MNIST dataset (Lecun et al., 1998) for training and evaluation, which is composed of 70,000 (60,000 for training and 10,000 for evaluation) 28×28 grayscale images. The SNNs were simulated using the Nengo python simulation tool (Bekolay et al., 2014), which provide numerical solutions to the differential equations of both LIF and ALIF neurons. The timestep for simulation was set to 5 ms, which is equal to the chosen refractory period for the neurons.

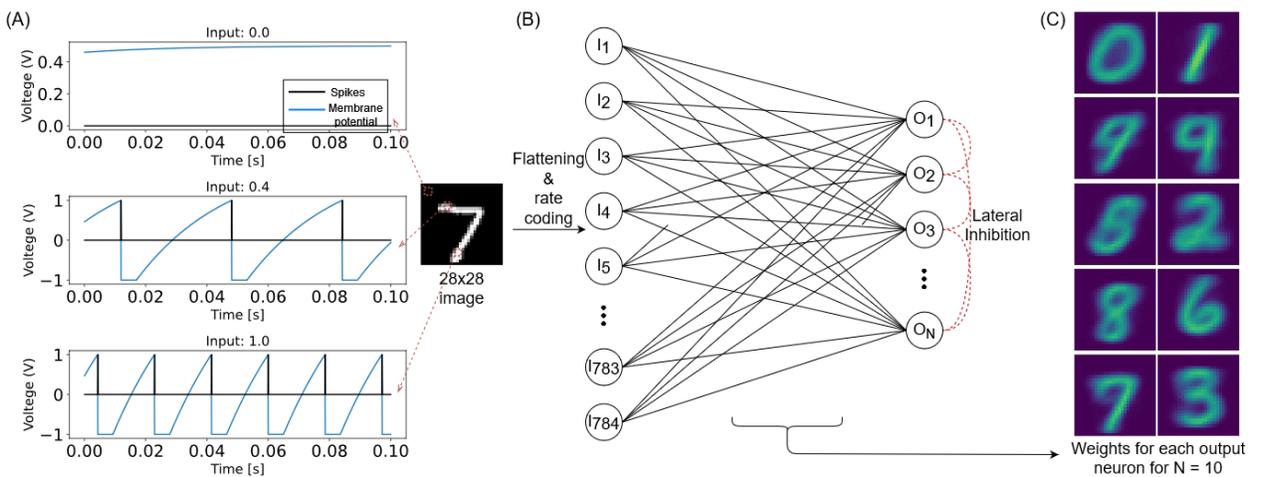

**Fig 3**: Representation of the SNN implementation used in this study to benchmark the VDSP learning rule with the MNIST classification task. (A) The response of the LIF neuron used in this study is plotted for input current of magnitude 0 (black pixel), 0.4 (grey pixel), and 1 (white pixel) for a duration of 100 ms. In (B), 28×28 grayscale image is rate encoded with the help of 784 input

LIF neurons. Each sample is presented for 350 ms. The input neurons are fully connected to the ALIF output neurons connected in Winner Takes All (WTA) topology for lateral inhibition. (C) The weight matrix for each of the ten output neurons.

The input layer is composed of 784 (28×28) LIF neurons (Fig. 3). The pixel intensity is encoded with frequency coding, where the spiking frequency of the neuron is proportional to the pixel value. It is essential, when using VDSP, to use different $v_{rest}$ and $v_{reset}$ values to discriminate inactive neurons and neurons that spiked recently (Fig. 2(C) and 2(D)). In our work, $v_{rest}$ is set to zero volt, and $v_{reset}$ is set to -1 V.

The output layer is modelled as ALIF neurons connected in a Winner Takes All (WTA) topology: on any output neuron spike occurrence, the membrane potential of all other neurons is clamped to zero for 10 ms. All the input neurons are connected to all the output neurons through synapses implementing the VDSP learning rule. The initial weights of these synapses were initialized randomly, with a uniform distribution between the minimum (0) and maximum (1) weight values. Each image from the MNIST database was presented for 350 ms with no wait time between images. The neuron parameters of input and output neurons used in this study are summarized in Table 1

**Table 1**: Parameters of LIF input and ALIF output neurons used in this study. In order to reproduce the results of this study, the same can be used in conjunction with proposed equations of the VDSP rule with a learning rate equal to $5x10^{-2}$.

| Property | Input layer | Output layer |
|---|---|---|
| Refractory period | 5 ms | 5 ms |
| Leak time constant | 30 ms | 30 ms |
| Reset voltage | -1 V | 0 V |
| Rest voltage | 0 V | 0 V |
| Threshold | 1 V | 1 V |
| Bias | 0.5 | 0 |
| Adaptation increment | - | 0.01 |
| Adaptation leak time constant | - | 1 s |
| WTA time constant | - | 10 ms |

Once trained, the weights were fixed, and the network was presented again with the samples from the training set, and all the output neurons were assigned a class based on activity during the presentation of digits of a different class. The 10,000 images from the test set of the MNIST database were presented to the trained network for testing the network. Based on the class of neuron with the highest number of spikes during sample presentation time, the predicted class was assigned. The accuracy was computed by comparing it with the true class. For larger networks, the cumulative spikes of all the neurons for a particular class were compared to evaluate the network's decision. The above could be easily realized in hardware with simple connections to the output layer neurons. More sophisticated machine learning classifiers like Support Vector Machines (SVMs) or another layer of spiking neurons can also be employed for readout to improve performance (Querlioz et al., 2012).

## Results and discussion

On training a network composed of 10 output neurons for a single epoch, with 60,000 training images of the MNIST database, we observe distinct receptive fields for all the ten digits (Fig. 3(C)). Note that the true labels are not used in the training procedure with the VDSP learning rule, and hence the learning is unsupervised. We report classification accuracy of $61.4 \pm 0.78\%$ ($Mean \pm S.D.$) based on results obtained from five different initial conditions.

### Presynaptic firing frequency dependence of VDSP

As stated previously, the VDSP rule does not use the presynaptic input current to compute $\Delta w$. Therefore, as the presynaptic input current changes, e.g., in between the samples of the MNIST dataset, the change in weight conductance, $\Delta w$, is affected. Figure 4 (A) presents the relation between the presynaptic firing frequency when the input current is changed and the $\Delta w = \text{VDSP}(\Delta t)$ window between two neurons with a fixed initial weight w=0.5. As the current gets larger, the presynaptic firing frequency is increased, and the window shortens. This has a normalizing effect on the learning mechanism of VDSP when subjected to different spiking frequency regimes.

In Figure 4 (B), we recreated a simplified version of the MNIST classification task using the WTA presented in the previous sections. Notably, there is no adaptation mechanism in the output layer, and the duration of the images is dynamically computed to have a maximum of ten spikes per pixel per image. These changes were made to specifically show the dependence of the input frequency on the accuracy, but they also affect the maximum reached accuracy in the case of VDSP. We ran the network with ten output neurons for one epoch with both VDSP and STDP with constant parameters. As expected, VDSP is much more resilient to the change in spiking input frequency. This effect is beneficial since the same learning rule can be used in hardware, and the learning can be accelerated by simply scaling the input currents. We note that neither the VDSP nor the STDP's parameters are maximized for absolute performance in this experiment, and we used the same weight normalizing function as (Diehl and Cook, 2015) for STDP.

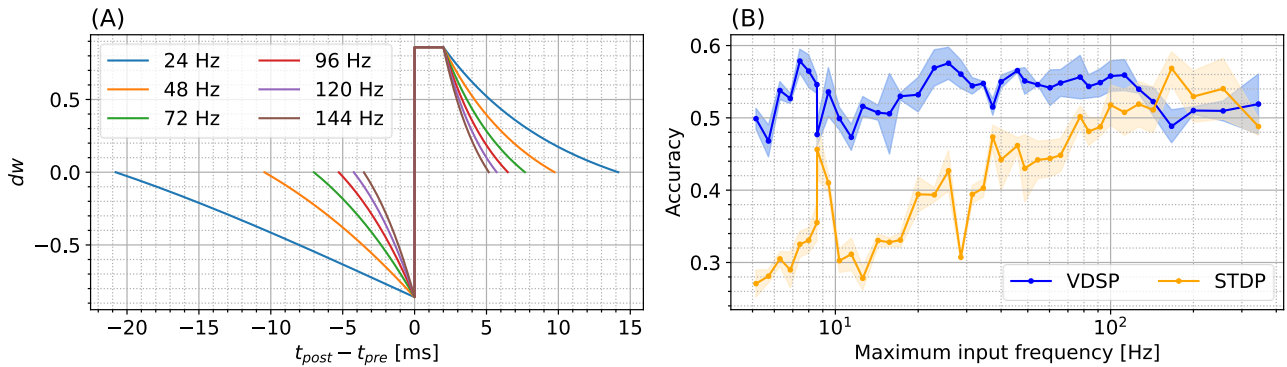

**Fig. 4**: Presynaptic firing frequency dependence of VDSP and STDP. Subfigure (A) shows the effect of scaling the presynaptic neuron input current on the VDSP update window for fixed weight $w = 0.5$ in a two neurons configuration. As the input current changes, the presynaptic neuron fires

at various frequencies indicated by the line colour. Higher presynaptic spiking frequencies result in smaller time windows. The plateau between $\Delta t \in [0, 2]$ ms is an artifact of the refractory period of 2 ms, where the membrane potential is kept at a reset value throughout. In (B), similar scaling is applied to the values of the pixels being fed to the presynaptic neurons during the MNIST classification task using the WTA architecture. Each point in (B) results from running the task 5 times with different random seeds using ten output neurons, with standard deviation shown with the light-coloured area under the curve. No adaptation mechanism was used for (B) to provide an unbiased comparison between classical STDP and VDSP in different spiking frequency regimes. No frequency-specific optimization was done during these experiments.

### Impact of network size and training time on VDSP

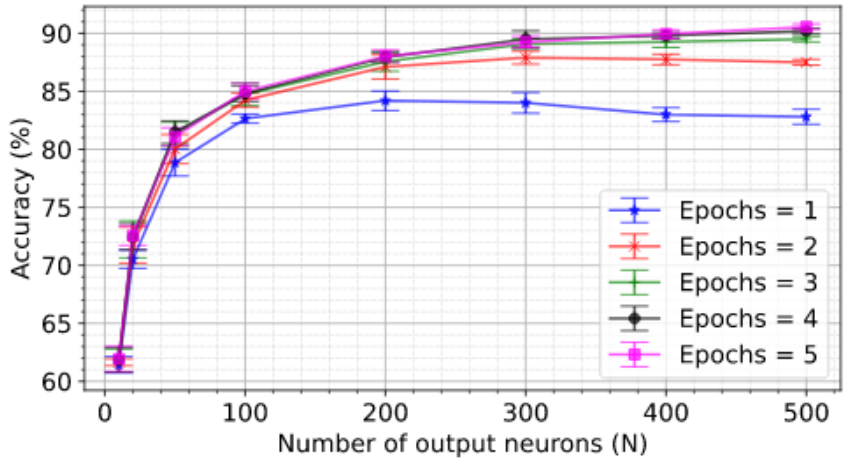

**Fig. 5**: A spiking neural network with 784 input neurons and N output neurons was trained on the training set (60,000 images) of the MNIST dataset for different numbers of epochs. The accuracy was computed on the test set (10,000) unseen images of the MNIST dataset. Networks with the number of output neurons ranging from 10 to 500 were trained for the number of epochs ranging from 1 to 5. Each experiment was conducted for five different initial conditions. The mean accuracy for five trials is plotted in the figure, with the error bar indicating the standard deviation.

To investigate the impact of the number of output neurons and epochs on classification accuracy, the two-layer network for MNIST classification is trained for up to five epochs and five hundred output neurons. The resulting accuracy for the different number of epochs and number of output neurons is shown in Fig. 5. Note that network hyperparameters were not re-optimized for these experiments (i.e., hyperparameters were optimized for a 50 output neuron topology only). Key performance numbers are tabulated in Table 2 and compared to the state-of-the-art accuracy reported in the literature. We observe equivalent or higher performance than the networks trained with the pair-based STDP in software simulations (Diehl and Cook, 2015) and hardware-aware simulations (Querlioz et al., 2013; Boybat et al., 2018; Guo et al., 2019) for most network sizes. This result validates the efficiency of the VDSP learning rule for solving computer vision pattern recognition tasks.

**Table 2**: State-of-the-art accuracy obtained with the STDP learning rule is tabulated for different numbers of epochs and output neurons. The performance achieved by training SNN with the VDSP rule is tabulated for various network sizes (number of output neurons) and epochs. Each experiment was repeated with five different initial conditions, and the accuracies are reported as (Mean±S.D.). Compared with the hardware-independent approach of pair based STDP (Diehl and Cook, 2015), we achieved 84.74 $\pm$ 1.08% for a network of 100 output neurons trained over three epochs. For a network of 400 output neurons trained over three epochs, we achieved 89.26 $\pm$ 0.54%.

| This work | | | Past studies | | | |
|---|---|---|---|---|---|---|
| **Neurons** | **Epochs** | **Accuracy (%) ($\mu \pm \sigma$)** | **Neurons** | **Epochs** | **Accuracy (%)** | **Ref.** |
| 10 | 1 | 61.4 ± 0.78 | 10 | 1 | 60 | (Querlioz et al., 2013) |
| 50 | 1 | 78.84 ± 1.28 | 50 | 1 | 76.8 | (Guo et al. 2019) |
| 50 | 3 | 81.3 ± 1.76 | 50 | 3 | 77.2 | (Boybat et al. 2018) |
| | | | 50 | 1 | 78.55 | (Demin, et al 2021) |
| | | | 50 | 3 | 81 | (Querlioz et al., 2013) |
| | | | 50 | - | 83.03 | (Oh, et al. 2019) |
| 100 | 3 | 84.74 ± 1.08 | 100 | 3 | 82.9 | (Diehl and Cook, 2015) |
| | | | 100 | 1 | 89.15 | (Demin, et al 2021) |
| | | | 200 | 17 | 91.63 | (Oh, et al. 2019) |
| 300 | 3 | 89.08 ± 0.49 | 300 | 3 | 93.5 | (Querlioz et al., 2013) |
| 400 | 3 | 89.26 ± 0.54 | 400 | 3 | 87 | (Diehl and Cook, 2015) |
| 500 | 5 | 90.56 ± 0.27 | | | | |

The performances of the network trained with VDSP are well aligned with hardware aware software simulations (Table 2) for simplified STDP and memristor simulation (Querlioz et al., 2011), resistive memory-based synapse simulation (Guo et al., 2019), PCM based synapse simulation (Boybat et al., 2018). VDSP has lower accuracies with respect to (Oh, et al. 2019) in their 50 and 200 neuron simulations, which can be explained by the different number of learning epoch and encoding strategy of the MNIST digits.

The comparable performance of VDSP with standard STDP can be attributed to the fact that the membrane potential is a good indicator of the history of input received by neurons and not just the last spike. In addition, the weight update in VDSP depends on the current weight, which regularizes the weight update and prevents the explosion or dying of weights. As in Supplementary Fig. S1, we observe a bimodal distribution of weights and clear receptive fields for a network of 50 output neurons. When this weight dependence is removed and clipping of weights between 0 and 1 is used, most weights become either zero or one, and receptive fields are not clear with current parameters (Supplementary Fig. S2).

### VDSP parameters optimization

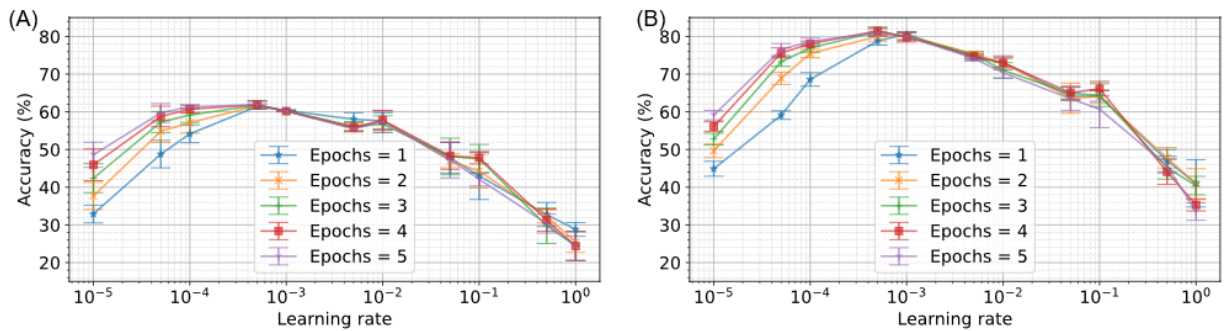

**Fig. 6**: Dependence of the performance on learning rate and number of epochs for different network sizes. In (A), a network with ten output neurons was trained on the MNIST dataset for different numbers of epochs and learning rates. Networks with learning rates ranging from $10^{-5}$ to 1 were trained for the number of epochs ranging from 1 to 5. Each experiment was conducted for five different initial conditions. The mean accuracy for five trials is plotted in the figure, with the error bar indicating the standard deviation. In (B), the experiments are repeated for 50 output neurons. As depicted, the optimum learning rate for a single epoch and ten neurons is $5 \times 10^{-4}$. Whereas, for 50 output neurons, the optimum learning rate for a single epoch is $10^{-3}$.

Convergence of the VDSP learning was possible with additional parameters optimization. Firstly, clear receptive fields require to decrease the weight of inactive pixels corresponding to the background. To penalize these background pixels, which do not contribute to the firing of the output neuron, we introduce a positive bias voltage in the input neurons of the MNIST classification SNN. This bias leads to a positive membrane potential of background neurons but does not induce firing. Consequently, the weight values are depressed according to the VDSP plasticity rule. Depressing the background neuron weight also balances the potentiation of foreground pixels and keeps in check the total weights contribution of an output neuron, thus preventing single neurons from always "winning" the competition. To validate the above hypothesis, we experimented training with zero bias voltage (Supplementary Fig. S3) and observed poor receptive fields.

The learning rate is a crucial parameter for regulating the granularity of weight updates. To study the impact of learning rate and the number of epochs on the performance, we train networks with learning rates ranging from $10^{-5}$ to 1 for up to five epochs. The resulting performance for five different runs is plotted for ten output neurons and 50 output neurons in Fig. 6. For a single epoch, we observe the optimal performance for ten output neurons at a learning rate of $5.10^{-3}$. For 50 output neurons and a single epoch, the optimal learning rate was $1.10^{-2}$. This result is indicative of the fact that the optimal learning rate increases for a greater number of neurons. Conventional STDP, on the other hand, has a minimum of two configurable parameters: learning rate and temporal sensitivity window for potentiation and depression. These are to be optimized to the dynamics of the input signal. VDSP has just one parameter and can be optimized based on the number of output neurons and training data size or the number of epochs, as discussed. There are many additional hyperparameters in a spiking neural network, such as time constant, thresholds, bias, and gain of the neurons, which can affect network performances. The neuron and simulation parameters tabulated in Table 1 were optimized with grid search performed on a network comprising 50 output neurons trained over a single epoch.

**Hardware choices for VDSP**

In the past, voltage dependent plasticity rules proposed triggering weight update on presynaptic neuron spike (Fusi et al., 2007; Diederich et al., 2018). Updating on presynaptic neuron spike is also an intuitive choice considering the forward directional computation graph for SNN. However, in the specific case of the output layer of multi-layer feedforward networks with WTA-based lateral inhibition, at most, one output neuron spikes at a time, and the output spike frequency would be significantly lower than the input spike frequency, reducing the frequency of weight updates required. Moreover, in multi-layer feedforward networks, activity in layers close to the output layer corresponds to the recognition of higher-level features and is a more attractive choice to synchronize the weight update. In addition, in networks for classification tasks, a convergence of layer size occurs from a large number of input neurons (for achieving high spatial resolution in neuromorphic sensors like DVS cameras, for instance) to a few neurons in the output layer. In hardware, a lower weight update frequency would imply lesser power consumption required in learning and a reduction in the learning time, thus providing greater flexibility with bandwidth available for inference.

The locality of the learning rule could be dependent on the hardware architecture. In the specific case of in-memory computing based neuromorphic hardware implementations, the synapse is physically connected to both postsynaptic and presynaptic neurons. State variables like the membrane potential of these neighboring neurons are readily available to the connecting synapse. Moreover, for memristive synapses, the dependence of weight change on initial weight is an inherent property of device switching. The proposed learning rule is attractive for implementing local learning in such systems.

For lateral inhibition in the output layer, the membrane potential of all the other output neurons is clamped to zero for 10 ms upon firing of any output neurons. This choice is inspired by the similar approach employed in (Querlioz et al., 2011; Oh, et al. 2019; Demin, et al. 2021). One alternative is using an equal number of inhibitory spiking neurons in the output layer (Diehl and Cook, 2015). However, using an equal number of inhibitory output neurons doubles the number of neurons, leading to the consumption of a significant silicon area when implemented on a neuromorphic chip. On the other hand, clamping the membrane potential does not require substantial circuit area and is a more viable option for hardware implementations.

We also evaluated the impact of injected Gaussian noise on neuron response for different input currents and noise distributions (supplementary information, figure S5). Gaussian noise centered around zero with different deviations was injected into the input neurons. While the membrane potential is substantially noisy in the case of mid-level noise injection, we do not observe a significant drop in performance. This feature makes VDSP an attractive choice of learning rule to be deployed on noisy analog circuits and nanodevices with high variability.

We also tested the applicability of the method for a network receiving random Poisson-sampled input spike patterns to drive the input layer. To elucidate this, a network of 10 output neurons was trained by feeding Poisson sampled spike trains to the input neuron with the frequency being proportional to the pixel value. The plots of membrane potential and neuron spike for different input values are presented in (supplementary information, figure S6(A-C)). The network was trained for one epoch and recognition accuracy of 58% was obtained on the test set. The resulting weight plots are shown in (supplementary information, figure S6(D)). Stable learning is observed and a small performance drop of 3% occurred as compared to constant input current.

### Conclusion and Future scope

In this work, we presented a novel learning rule for unsupervised learning in SNNs. VDSP is solving some of the limitations of STDP for future deployment of unsupervised learning in SNN. Firstly, as plasticity is derived from the membrane potential of the pre-synaptic neuron, VDSP on hardware would reduce memory requirement for storing spike traces for STDP based learning. Hence, larger and more complex networks can be deployed on neuromorphic hardware. Secondly, we observe that the temporal window adapts to the input spike frequencies. This property solves the complexity of STDP implementation, which requires STDP time window adjustment to the spiking frequency. This intrinsic time window adjustment of VDSP could be exploited to build hierarchical neural networks with adaptive temporal receptive fields (Paredes-Valles, Scheper and Croon de, 2020; Maes, Barahona and Clopath 2021). Thirdly, the frequency of weight update is significantly lower than the STDP, as we do not perform weight updates on both presynaptic and postsynaptic neuron spike events. This decrease in weight updates frequency by a factor of two is of direct interest for increasing the learning speed of SNN simulation and operation. Furthermore, this improvement is obtained without trading off classification performances on the MNIST dataset, thus validating the applicability of VDSP rule in pattern recognition. The impact of hyperparameters (learning rate, network size, and the number of epochs) is discussed in detail with the help of simulation results.

In the future, we will investigate the implementation of VDSP in neuromorphic hardware based on emerging memories. Also, future work should consider investigating the proposed learning rule for multi-layer feed-forward networks and advanced network topologies like Convolutional Neural Networks (CNNs) (Lee et al., 2018; Kheradpisheh et al., 2018) and Recurrent Neural Networks (RNNs) (Gilson, 2010). Finally, using this unsupervised learning rule in conjunction with gradient-based supervised learning is an appealing aspect to be explored in future works.

### Conflict of Interest

*The authors declare that the research was conducted in the absence of any commercial or financial relationships that could be construed as a potential conflict of interest.*

### Author Contributions

J.R., Y.B., F.A., J.M.P and D.D. contributed to formulating the study. N.G., I.B. designed and performed the experiments and derived the models. N.G., I.B., F.A., and Y.B. analyzed the data.

T.C.S. contributed to realizing the plasticity rule in Nengo. All authors provided critical feedback and helped shape the research, analysis, and manuscript.

**Funding**

We acknowledge financial supports from the EU: ERC-2017-COG project IONOS (# GA 773228) and CHIST-ERA UNICO project. This work was also supported by the Natural Sciences and Engineering Research Council of Canada (NSERC) [funding reference number 559730] and Fond de Recherche du Québec Nature et Technologies (FRQNT).

**Contribution statement**

With the advent of IoT devices and big data, a massive amount of data is available to train machine learning algorithms. However, most of the data is unlabeled and cannot be processed by state-of-the-art supervised learning rules. We propose a novel brain-inspired unsupervised learning. While the present Hebbian-based unsupervised learning rule (STDP) is efficient in terms of performance, deployment of the same on neuromorphic hardware is expensive as it requires area-consuming capacitors to store precise spike times and requires significant peripheral circuitry. The proposed learning rule is efficient to be deployed on such hardware and in performing pattern recognition tasks. We show with rigorous mathematical analysis and simulations that the proposed learning rule is in line with Hebb's plasticity principles. We achieve greater than 90% accuracy on the handwritten digit recognition task, and the network performance is robust against injected noise. This learning rule is hence a suitable option for deployment on both analog and digital neuromorphic hardware and should improve the performance of state-of-the-art SNN topologies.